\documentclass[sigconf]{acmart}

\usepackage{booktabs} 
\usepackage{graphicx}
\usepackage{amsmath}
\usepackage{booktabs}
\usepackage{multirow}
\usepackage{colortbl}
\usepackage{makecell}
\usepackage{diagbox}

\setcopyright{acmcopyright}

\acmConference[SA '22 Conference Papers]{SIGGRAPH Asia 2022 Conference
Papers}{December 6--9, 2022}{Daegu, Republic of Korea}
\acmBooktitle{SIGGRAPH Asia 2022 Conference Papers (SA '22 Conference
Papers), December 6--9, 2022, Daegu, Republic of
Korea}
\acmDOI{10.1145/3550469.3555376}
\acmISBN{978-1-4503-9470-3/22/12}

\usepackage{xcolor}
\definecolor{myred}{rgb}{0.8,0,0}
\definecolor{mygreen}{rgb}{0,0.8,0}
\usepackage{pifont}
\newcommand{\cmark}{{\color{mygreen}\ding{51}}}

\usepackage[ruled]{algorithm2e} 

\SetAlFnt{\small}
\SetAlCapFnt{\small}
\SetAlCapNameFnt{\small}
\SetAlCapHSkip{0pt}

\begin{document}
\title{Efficient Neural Radiance Fields for Interactive Free-viewpoint Video}

\author{Haotong Lin}
\authornote{Both authors contributed equally to this research.}
\author{Sida Peng}
\authornotemark[1]
\email{haotongl@zju.edu.cn}
\email{pengsida@zju.edu.cn}
\affiliation{%
 \institution{State Key Laboratory of CAD\&CG, Zhejiang University}
 \country{China}
 }

\author{Zhen Xu}
\author{Yunzhi Yan}
\author{Qing Shuai}
\email{zhenx@zju.edu.cn}
\email{yanyz@zju.edu.cn}
\email{s_q@zju.edu.cn}
\affiliation{%
 \institution{Zhejiang University, China}
 }

\author{Hujun Bao}
\author{Xiaowei Zhou}
\authornote{Corresponding author.}
\email{bao@cad.zju.edu.cn}
\email{xwzhou@zju.edu.cn}
\affiliation{%
 \institution{State Key Laboratory of CAD\&CG, Zhejiang University}
 \country{China}
 }

\begin{abstract}
This paper aims to tackle the challenge of efficiently producing interactive free-viewpoint videos. 
Some recent works equip neural radiance fields with image encoders, enabling them to generalize across scenes. When processing dynamic scenes, they can simply treat each video frame as an individual scene and perform novel view synthesis to generate free-viewpoint videos. However, their rendering process is slow and cannot support interactive applications. 
A major factor is that they sample lots of points in empty space when inferring radiance fields. 
We propose a novel scene representation, called ENeRF, for the fast creation of interactive free-viewpoint videos. Specifically, given multi-view images at one frame, we first build the cascade cost volume to predict the coarse geometry of the scene. The coarse geometry allows us to sample few points near the scene surface, thereby significantly improving the rendering speed. This process is fully differentiable, enabling us to jointly learn the depth prediction and radiance field networks from RGB images. 
Experiments on multiple benchmarks show that our approach exhibits competitive performance while being at least 60 times faster than previous generalizable radiance field methods. 
\end{abstract}



\maketitle

\begin{figure}[h]
    \centering
    \includegraphics[width=\linewidth]{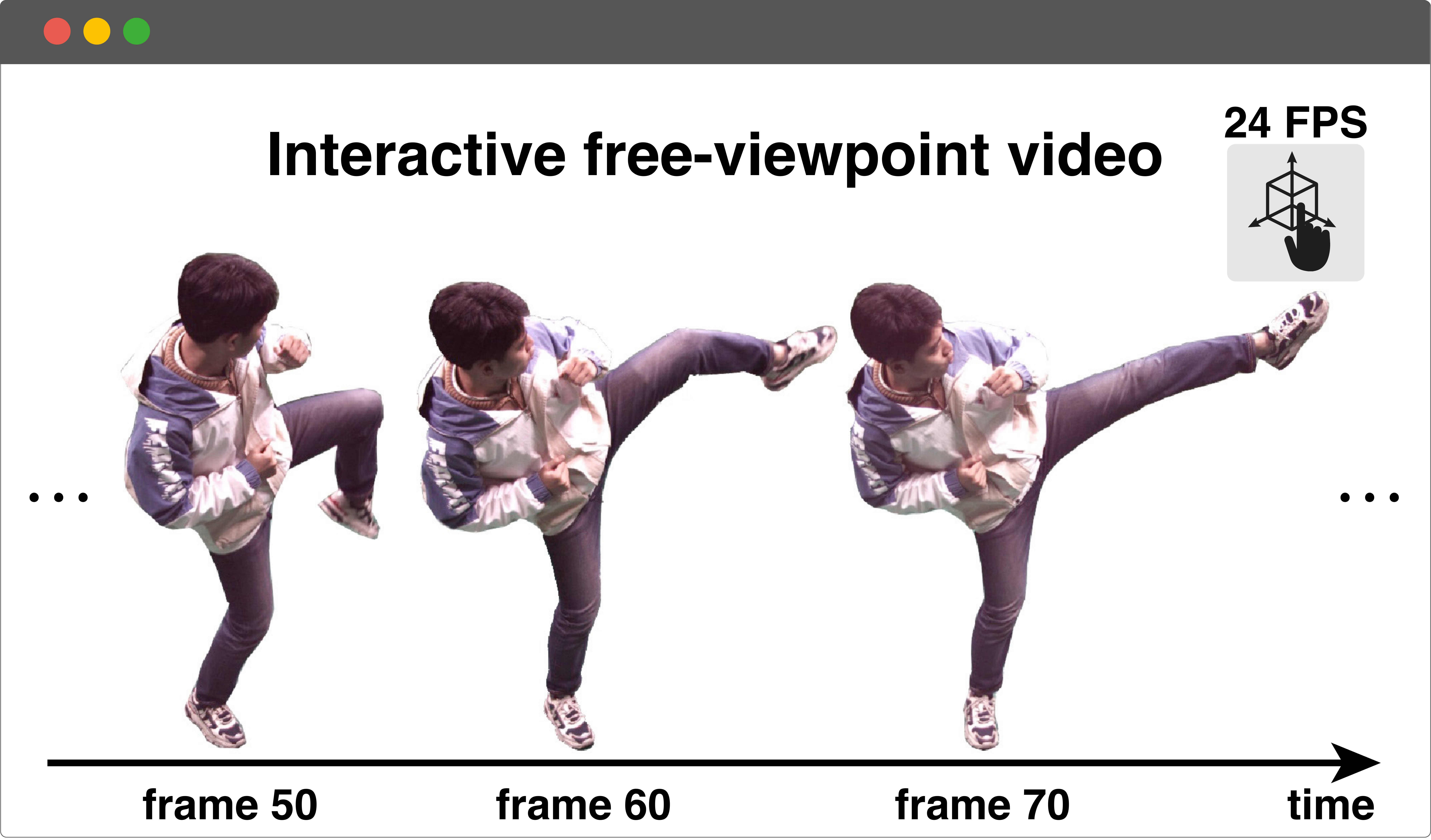}
    \caption{{
    The proposed method achieves photorealistic view synthesis for dynamic scenes at interactive frame rates. 
    Please refer to our video for the real-time demos that render free-viewpoint videos of dynamic scenes.}
    }
    \label{fig:teaser}
\end{figure}

\section{Introduction}


Free-viewpoint videos have a variety of applications such as virtual tourism, immersive telepresence, and movie production. These applications generally require us to accomplish two things. First, given multi-view videos, we need to quickly generate the free-viewpoint video, which is essential for immersive telepresence. Second, rendering novel views of target scenes should be fast, which is important for user experience. Neural radiance field (NeRF) \cite{mildenhall2020nerf} represents scenes as density and radiance fields, enabling it to achieve photorealistic novel view synthesis. \cite{pumarola2020d, park2021nerfies} leverage deformation fields to extend NeRF to dynamic scenes and thus can produce free-viewpoint videos. However, these methods generally require a long training process for each new scene. Moreover, their rendering speed is slow.


Recently, \cite{yu2021plenoctrees, garbin2021fastnerf} propose to cache radiance fields into highly efficient data structures, which significantly improves the rendering speed. To handle dynamic scenes, FastNeRF \cite{garbin2021fastnerf} augments the cached scene with deformation fields that transform the world-space points to the space of the cached scene. Although it can render free-viewpoint videos in real-time, there are two problems. First, the deformation fields are difficult to be recovered from videos when there exist complex scene motions, as discussed in \cite{peng2021animatable}. Second, it still needs a lengthy optimization process, making it impossible to quickly produce free-viewpoint videos from new data.


Another line of works \cite{yu2020pixelnerf, wang2021ibrnet} train networks to predict radiance fields from multi-view images and can generalize to unseen scenes. These methods overcome the two problems of FastNeRF \cite{garbin2021fastnerf}.
First, when processing dynamic scenes, they are able to simply treat every frame as an individual scene and perform novel view synthesis, instead of recovering temporal motions like \cite{pumarola2020d, garbin2021fastnerf}.
Second, thanks to the pretraining, they can be quickly fine-tuned to produce photorealistic rendering results on new scenes. However, their rendering speed is slow, as they require many forward passes through a neural network for rendering a pixel.


In this paper, we propose a novel rendering approach, called ENeRF, for efficiently generating interactive free-viewpoint videos. Our innovation lies in introducing a learned depth-guided sampling strategy that greatly speeds up the rendering of generalizable radiance field methods. Specifically, for the target view, we construct the cascade cost volume, which is used to predict a depth probability distribution. The depth probability distribution gives an interval in which the surface may be located. With the depth interval, we only need to sample few 3D points along the ray and thus improve the rendering speed of previous methods~\cite{wang2021ibrnet,chen2021mvsnerf}. Similar to \cite{chen2021mvsnerf}, we predict radiance fields using the cost volume feature, which makes our method generalize well to new scenes. Moreover, the whole pipeline is fully differentiable, so the depth-guided sampling strategy can be jointly learned with NeRF from only multi-view RGB images.



We evaluate our approach on the DTU~\cite{jensen2014large}, Real Forward-facing~\cite{mildenhall2020nerf,mildenhall2019local}, NeRF Synthetic~\cite{mildenhall2020nerf}, DynamicCap~\cite{habermann2021} and ZJU-MoCap~\cite{peng2021neural} datasets, which are widely-used benchmark datasets for novel view synthesis. Our approach makes significant rendering acceleration while achieving competitive results with baselines across all datasets. Specifically, our approach runs at least 60 times faster than previous generalizable radiance field methods~\cite{yu2020pixelnerf,wang2021ibrnet,chen2021mvsnerf}.  
We also show that our approach can produce reasonable depth maps by supervising the networks with only images.

In summary, this work has the following contributions: 1) We propose a novel approach that utilizes a learned depth-guided sampling strategy to improve the rendering efficiency of generalizable radiance field methods. 2) We show that the depth-guided sampling can be jointly learned with NeRF from only RGB images. 3) We demonstrate that our approach runs significantly faster than previous methods while being competitive on the rendering quality on several view synthesis benchmarks. 4) We demonstrate the capability of our method to synthesize novel views of human performers in real-time.  Our code is available at \href{https://zju3dv.github.io/enerf/}{https://zju3dv.github.io/enerf}.

\section{Related work}

\paragraph{Novel view synthesis.}
Some methods achieves the free-viewpoint rendering based on the light field interpolation \cite{gortler1996lumigraph, levoy1996light, davis2012unstructured} or image-based rendering \cite{zitnick2004high, chaurasia2013depth,flynn2016deepstereo, kalantari2016learning, penner2017soft, Riegler2020FVS}.
Recently, neural representations \cite{sitzmann2019deepvoxels, liu2019neural, li2020crowdsampling, lombardi2019neural, shih20203d,liu2019learning,jiang2020sdfdiff, wizadwongsa2021nex} are widely used for novel view synthesis, which can be optimized from input images and achieve photorealistic rendering results.
Neural radiance field (NeRF) \cite{mildenhall2020nerf} represents scenes as continuous color and density fields and yields impressive rendering results.
There are some works \cite{yu2020pixelnerf, trevithick2020grf, wang2021ibrnet, chen2021mvsnerf, liu2021neural, johari2021geonerf} that attempt to improve NeRF on the training speed.
They use 2D CNNs to process input images and decode multi-view features to target radiance fields. By pre-training networks, they can be quickly fine-tuned to produce high-quality rendering results on new scenes.
More recently, \cite{yu2021plenoxels, muller2022instant, sun2021direct, chen2022tensorf, sun2022neuconw} develop hybrid or explicit structures based on NeRF and achieve super-fast convergence for radiance fields reconstruction.
Another line of works \cite{liu2020neural, yu2021plenoctrees, reiser2021kilonerf, garbin2021fastnerf, hedman2021baking} aim to accelerate the rendering speed of NeRF.
By caching neural radiance fields, \cite{garbin2021fastnerf} can synthesize photorealistic images in real time.
Some methods \cite{fang2021neusample, arandjelovic2021nerf, barron2021mip, piala2021terminerf, neff2021donerf} improve the sampling strategy of NeRF to accelerate the rendering.
DONeRF~\cite{neff2021donerf} uses the depth-guided sampling for NeRF. However, it needs depth supervision and requires per-scene optimization.
There exists some methods~\cite{deng2022depth,roessle2022dense,rematas2022urban,wei2021nerfingmvs} exploring depth supervision to facilitate the reconstruction of neural radiance fields.
In the field of 3D generation, \cite{gu2021stylenerf} accelerates the rendering of 3D GAN with an upsampler to upsample a low-resolution feature map. ~\cite{chan2022efficient} enables real-time rendering of 3D GAN using a tri-plane representation. The above works mainly focus on view synthesis of static scenes.


\begin{figure*}[t]
    \centering
    \includegraphics[width=0.95\textwidth]{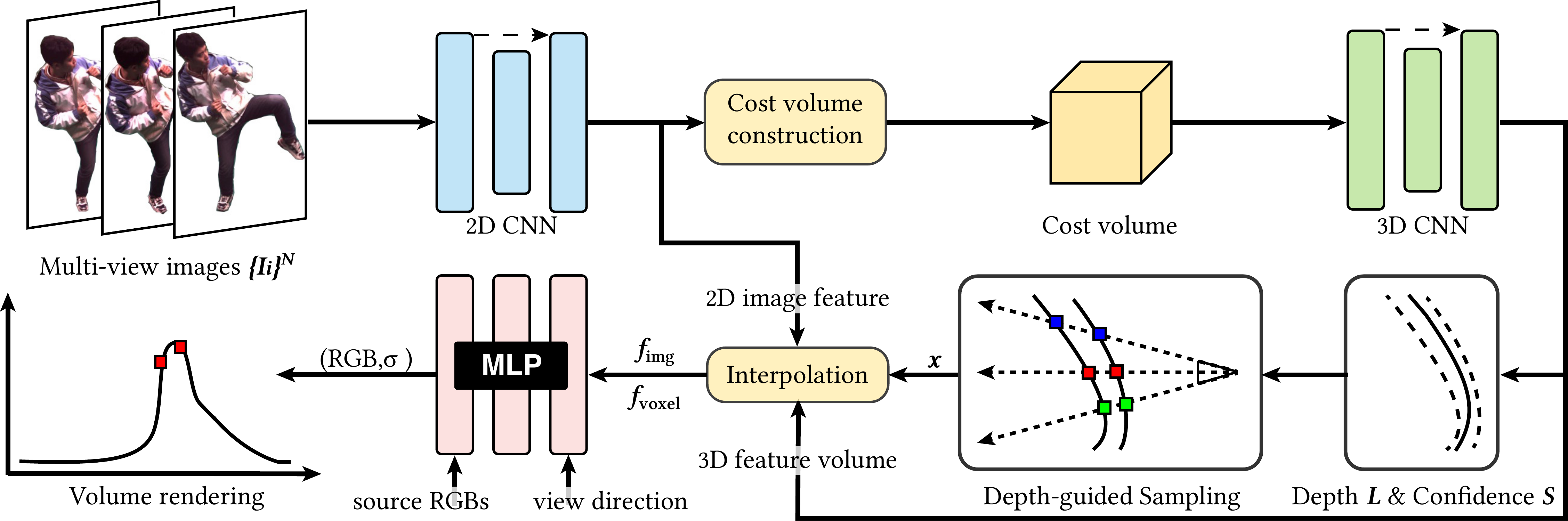}
    \caption{\textbf{Illustration of the proposed approach.} Given multi-view images of a static scene or a dynamic scene at one frame, we first construct the cascade cost volume, which is processed to output the 3D feature volume and the coarse scene geometry (represented by depth and confidence maps). 
    The estimated geometry guides us to sample around the surface, which significantly accelerates the volume rendering process.
    Also, the 3D feature volume  provides rich geometry-aware information for generalizable radiance fields construction.
    All network components are trained end-to-end using only RGB images. }
    \label{fig:framework}
\end{figure*}

\paragraph{View synthesis of dynamic scenes.} 
Some methods \cite{li2020neural, xian2021space,li2021neural3d} treat dynamic scenes as 4D domain and add the time dimension to the input spatial coordinate to implement the space-time radiance fields.
To produce free-viewpoint videos from monocular videos, \cite{park2021nerfies, pumarola2020d, park2021hypernerf} propose to aggregate the temporal information of input videos.
DNeRF \cite{pumarola2020d} augments NeRF with deformation fields that establish correspondences between the canonical scene and the scenes at different time instants.
\cite{zhang2021editable} develop a layered neural representation based on DNeRF, thereby better handling dynamic scenes containing multiple persons and static background.
Although these methods achieve impressive rendering results on dynamic scenes, they typically have a slow rendering speed.
FastNeRF \cite{garbin2021fastnerf} caches the canonical NeRF and uses a small MLP to predict the deformation fields, which improves the rendering speed. However, it can only achieve the real-time rendering on $300 \times 300$ images. Moreover, the deformation fields are difficult to estimated from videos on complex dynamic scenes, as discussed in Peng et al. \shortcite{peng2021animatable}.
More recently, \cite{zhang2022neuvv,wang2022fourier} introduce 4D volumetric representations, which achieves efficient generation of free-viewpoint videos.
They generally handle videos with short frames.
In the field of human rendering, \cite{peng2021neural, peng2021animatable, weng2022humannerf, lombardi2021mixture,liu2021neuralactor} utilize human priors to model human motions and thus accomplish the free-viewpoint rendering of dynamic human performers.
Generalizable radiance fields \cite{wang2021ibrnet, yu2020pixelnerf} can be used to render dynamic scenes by individually processing the scene at each time, which overcomes the problem of recovering temporal motions. However, its rendering speed is slow.

\paragraph{Multi-view stereo methods.}
The cost volume has been widely used for depth estimation in multi-view stereo (MVS) methods. 
MVSNet~\cite{yao2018mvsnet} proposes to build the 3D cost volume from 2D image features and regularizes it with a 3D CNN. 
This design enables end-to-end training of the network with ground truth depth and achieves impressive results.
However, memory consumption of MVSNet is huge.
To overcome this problem, following works improve it with recurrent plane sweeping~\cite{yao2019recurrent} or coarse-to-fine architectures~\cite{chen2019point,yu2020fast,gu2020cascade,cheng2020deep}.
In the field of novel view synthesis, \cite{chen2021mvsnerf, chibane2021stereo} attempt to combine MVS methods with neural radiance field methods.
They~\shortcite{chen2021mvsnerf} aim to facilitate the construction of the radiance fields leveraging stereo features. In contrast, our method focuses on accelerating the rendering process with explicit geometry. Another line of works \cite{oechsle2021unisurf,yariv2021volume,wang2021neus} combine multi-view stereo and neural radiance fields to facilitate geometry reconstruction.

\section{Method}
Given multi-view images, our task is to generate images of novel views at interactive frame rates. 
As shown in Figure~\ref{fig:framework}, to generate a novel view from multi-view images, we first extract multi-scale feature maps from input images, which are used to estimate coarse scene geometry and neural radiance fields (Sec.~\ref{sec:feature_extraction}).
Given extracted feature maps, we construct the cascade cost volume to obtain the 3D feature volume and the coarse scene geometry (Sec.~\ref{sec:depth_prediction}).
The feature volume provides geometry-aware information for radiance fields construction, and the estimated coarse geometry allows us to sample around the surface, which significantly accelerates the rendering process (Sec.~\ref{sec:nerf_construction}).
All networks are trained end-to-end with the view synthesis loss using only RGB images (Sec.~\ref{sec:training}).

\subsection{Multi-scale image feature extraction}
\label{sec:feature_extraction}
To build the cascade cost volume, we first extract multi-scale image features from input views $\{I_i\}_{i=1}^{N}$ using a 2D UNet. 
As shown in Figure~\ref{fig:framework},
we first feed an input image $I_i \in \mathbb{R}^{H \times W \times 3}$ into the encoder to obtain low-resolution feature maps $F_{i, 1} \in \mathbb{R}^{H/4 \times W/4 \times 32}$. 
Then we use two deconvolution layers to upsample the feature maps and obtain the other two-stage feature maps $F_{i, 2}  \in \mathbb{R}^{H/2 \times W/2 \times 16}$ and $F_{i, 3}  \in \mathbb{R}^{H \times W \times 8}$  . 
 $F_{i, 1}$ and  $F_{i, 2}$ are used to construct cost volumes, and  $F_{i, 3}$ is used to reconstruct neural radiance fields.



\subsection{Coarse-to-fine depth prediction}
\label{sec:depth_prediction}

To efficiently obtain the coarse scene geometry (represented by a high-resolution depth map) from multi-view images, 
we construct the cascade cost volume under the novel view frustum to predict the depth for the novel view.
Specifically, we first construct a coarse-level low-resolution cost volume and recover a low-resolution depth map from this cost volume. 
Then we construct a fine-level high-resolution cost volume utilizing the depth map estimated in the last step.
The fine-level cost volume is processed to produce a high-resolution depth map and a 3D feature volume.

\paragraph{Coarse-level cost volume construction.}
Given initial scene depth range, we first sample a set of depth planes $\{\mathbf{L}_{i} | i = 1, ..., D\}$.
Following learning-based MVS methods~\cite{yao2018mvsnet}, we construct the cost volume by warping image features $F_{i, 1}$ into $D$ sweeping planes. 
Given camera intrinsic, rotation and translation $[\mathbf{K}_i, \mathbf{R}_i, \mathbf{t}_i]$ of input view $I_i$ and $[\mathbf{K}_t, \mathbf{R}_t, \mathbf{t}_t]$ of target view, 
the homography warping is defined as:
\begin{equation}
    \mathbf{H}_i(z) = \mathbf{K}_i \mathbf{R}_i \Big(\mathbf{I} + \frac{(\mathbf{R}_i^{-1} \mathbf{t}_i - \mathbf{R}_t^{-1} \mathbf{t}_t) \mathbf{a}^T \mathbf{R}_t}{z} \Big) \mathbf{R}_t^{-1} \mathbf{K}_t^{-1},
\end{equation}
where $\mathbf{a}$ denotes the principal axis of the target view camera, $\mathbf{I}$ is the identity matrix and $\mathbf{H}_i(z)$ warps a pixel $(u, v)$ in the target view at depth $z$ to the input view. 
The warped feature maps $F^w_{i} \in \mathbb{R}^{D \times H/8 \times W/8 \times 32}$ are defined as:
\begin{equation}
    F^w_{i}(u, v, z) = F_{i, 1}(\mathbf{H}_i(z)[u, v, 1]^T).
    \label{equ:warping}
\end{equation}
Based on the warped feature maps, we construct the cost volume by computing the variance of multi-view features $\{F^w_{i}(u, v, z) | i=1, ..., N\}$ for each voxel. 

\paragraph{Depth probability distribution.}
\label{depth_prob}
Given the constructed cost volume, we use a 3D CNN to process it into a depth probability volume $\mathbf{P} \in \mathbb{R}^{D \times H/8 \times W/8}$.
Similar to \cite{cheng2020deep}, we compute a depth distribution based on the depth probability volume. 
For a pixel $(u, v)$ in the target view, we can obtain its probability at a certain depth plane $\mathbf{L}_{i}$ by linearly interpolating the depth probability volume, which is denoted as $\mathbf{P}_{i}(u, v)$. 
Then the depth value of pixel $(u, v)$ is defined as the mean $\hat{\mathbf{L}}$ of the depth probability distribution:
\begin{equation}
    \hat{\mathbf{L}}(u, v) = \sum_{i = 1}^{D} \mathbf{P}_{i}(u, v) \mathbf{L}_{i}(u, v),
\end{equation}
and its confidence is defined as the standard deviation $\hat{\mathbf{S}}$:
\begin{equation}
    \hat{\mathbf{S}}(u, v) = \sqrt{\sum_{i = 1}^{D} \mathbf{P}_{i}(u, v) (\mathbf{L}_{i}(u, v) - \hat{\mathbf{L}}(u, v))^2}.
\end{equation}

\paragraph{Fine-level cost volume construction and processing.}
The depth probability distribution with the mean $\hat{\mathbf{L}}(u, v)$ and the standard deviation $\hat{\mathbf{S}}(u, v)$ determines where the surface may be located.
Specifically, the surface should be located in the depth range defined as follows,  
\begin{equation}
    \hat{\mathbf{U}}(u, v) =  
    [\hat{\mathbf{L}}(u, v) - \lambda \hat{\mathbf{S}}(u, v), \hat{\mathbf{L}}(u, v) + \lambda \hat{\mathbf{S}}(u, v)],
\end{equation}
where $\lambda$ is a hyper-parameter that  determines how large the depth range is.
We simply set $\lambda$ to $1$.
To construct the fine-level cost volume, we first upsample the estimated depth range map $\hat{\mathbf{U}} \in \mathbb{R}^{H/8 \times W/8 \times 2}$ four times.
Given the depth range map, we uniformly sample $D'$ depth planes within it and construct the fine-level cost volume by applying the warping to the feature maps $F_{i, 2}$ similar to the Equation~\eqref{equ:warping}.
Then we use a 3D CNN to process this cost volume to obtain a depth probability volume and a 3D feature volume. 
Following the processing step of the coarse-level depth probability volume, we get a finer depth range map $\hat{\mathbf{U}}' \in \mathbb{R}^{H/2 \times W/2 \times 2}$, which will guide the sampling for volume rendering.

\subsection{Neural radiance fields construction}
\label{sec:nerf_construction}

NeRF \cite{mildenhall2020nerf} represents the scene as color and volume density fields. To generalize across scenes, similar to \cite{yu2020pixelnerf, wang2021ibrnet, chen2021mvsnerf}, our method assigns features to arbitrary point in 3D space. Inspired by PixelNeRF \cite{yu2020pixelnerf}, for any 3D point, we project it into input images and then extract corresponding pixel-aligned features from $\{F_{i, 3}|i=1, ..., N\}$, which are denoted as $\{f_i | i=1, ..., N\}$, where $N$ is the number of input views. 
These features are then aggregated with a pooling operator $\psi$ to output the final feature $f_{\text{img}} = \psi(f_1, ..., f_N)$.
The design of this pooling operator $\psi$ is borrowed from IBRNet~\cite{wang2021ibrnet} and its details is described in the supplementary material.
To leverage the geometry-aware information provided by the MVS framework, we also extract the voxel-aligned feature from the 3D feature volume by transforming the 3D point into the query view and trilinearly interpolating this 3D feature volume to get the voxel feature, which is denoted as $f_{\text{voxel}}$. Our method passes $f_{\text{img}}$ and $f_{\text{voxel}}$ into an MLP network to obtain the point feature and density, which is defined as:
\begin{equation}
    f_p, \sigma =\phi(f_{\text{img}}, f_{\text{voxel}}),
    \label{eq:density}
\end{equation}
where $\phi$ denotes the MLP network.
We estimate the color $\hat{\mathbf{c}_p}$ of this 3D point viewed in direction $\mathbf{d}$ by predicting blending weights for the image colors $\{\mathbf{c}_i\}^{N}_{i=1}$ in the source views.
Specifically, we concatenate the point feature $f_p$ with the image feature $f_i$ and $\Delta \mathbf{d}_i$, and feed them into an MLP network to yield the blending weight $w_i$ defined as:
\begin{equation}
    w_i = \varphi (f_p, f_i, \Delta \mathbf{d}_i),
\end{equation}
where $\varphi$ denotes the MLP network and $\Delta \mathbf{d}_i$ is defined as the concatenation of the norm and direction of $\mathbf{d}_i - \mathbf{d}$.
$\mathbf{d}$ and $\mathbf{d}_i $ are the ray directions of the 3D point under the target view and corresponding source view, respectively. 
The color $\hat{\mathbf{c}_p}$  is blended via a soft-argmax operator as the following,
\begin{equation}
    \hat{\mathbf{c}}_p = \sum^N_{i=1}\frac{ \text{exp}(w_i) \mathbf{c}_i}{\sum^N_{j=1} \text{exp}(w_j)}. 
    \label{eq:color}
\end{equation}

\paragraph{Depth-guided sampling for volume rendering.} Given a viewpoint, our method renders the radiance field into an image with the volume rendering technique \cite{mildenhall2020nerf}. Consider pixel $(u, v)$. We have its depth range $\hat{\mathbf{U}}'(u, v)$ estimated from depth prediction module (Sec.~\ref{sec:depth_prediction}). Then we uniformly sample $N_k$ points $\{\mathbf{x}_k | k = 1, ..., N_k\}$ within this depth range. Finally, our method predicts the densities and colors for these points based on Equations~\eqref{eq:density} and ~\eqref{eq:color}, which are accumulated into the pixel color.

\subsection{Training}
\label{sec:training}

During training, the gradients are back-propagated to the estimated depth probability distribution through sampled 3D points, so that the depth probability volume can be jointly learned with neural radiance fields from only images. 
Following \cite{mildenhall2020nerf}, we optimize our model with the mean squared error that measures the difference between the rendered and ground-truth pixel colors. The corresponding loss is defined as:
\begin{equation}
    \mathcal{L}_{mse} = \frac{1}{N_r}\sum_{i=1}^{N_r}\|\hat{\mathbf{C}}_i - \mathbf{C}_i\|^2_2,
\end{equation}
where $N_r$ is the number of sampled rays at each iteration and $\hat{\mathbf{C}}_i$ and $\mathbf{C}_i$ are the rendered and ground-truth color, respectively.
Thanks to the depth-guided sampling, we sample only few points and spend little GPU memory during training. Therefore, we can also sample image patches and supervise them using the perceptual loss~\cite{johnson2016perceptual}. 
\begin{equation}
    \mathcal{L}_{perc} =  \frac{1}{N_i}\sum_{i=1}^{N_i}\| \ \Phi ( \hat{\mathbf{I}}_i) - \Phi( \mathbf{I}_i)\|,
\end{equation}
where $N_i$ is the number of image patches and $\Phi$ is the definition of perceptual function (a VGG16 network).
The final loss function is:
\begin{equation}
    \mathcal{L} =  \mathcal{L}_{mse} + \lambda' \mathcal{L}_{perc}.
\end{equation}
In practice, we set $\lambda' = 0.01$ in all experiments.

\begin{figure}[t]
    \centering
    \includegraphics[width=0.48\textwidth]{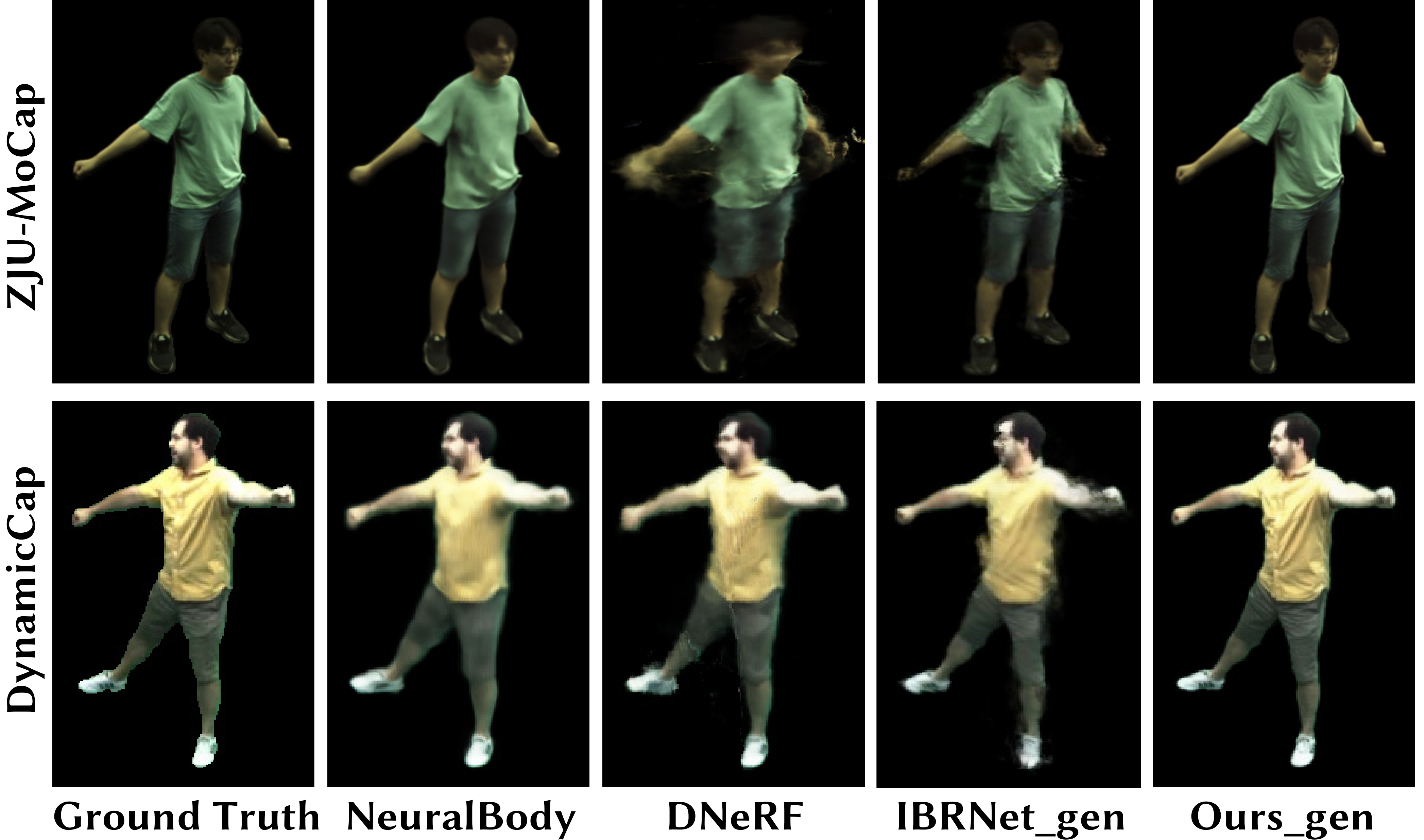}
    \caption{\textbf{Quantitative results on dynamic scenes.} ``*\_gen'' means that these models are directly applied to input images without being additionally fine-tuned on input videos.}
    \label{fig:dynamiccap}
\end{figure}


\subsection{Implementation details}
Our generalizable rendering model is trained on an RTX 3090 GPU using Adam~\cite{kingma2014adam} optimizer with an initial learning rate of $5e^{-4}$ and we half the learning rate every 50k iterations.
The model tends to converge after about 200k iterations and it takes about 15 hours.
Given an unseen scene, we can finetune our pre-trained model on this scene.
Fine-tuning on the new scene generally takes from 10 minutes to 2 hours on an RTX 3090 GPU depending on the number of images.
In practice, we sample 64 and 8 depth planes for the coarse-level and fine-level cost volumes, respectively.
The proposed rendering model takes from 2 to 3 input source views depending on the number of available cameras.
In practice, we set the number of input source views to 2 on dynamic scenes (sparse cameras) and 3 on static scenes (dense cameras) in our experiments.
The number of samples per ray is set to 2. We add the sensitivity analysis of number of input views and samples per ray in the supplementary material.
The depth ranges for the cost volume is obtained from the structure from motion~\cite{schonberger2016structure} algorithm. 
Please refer to the supplementary material for network structures and other implementation details.

\section{Experiments}

\subsection{Experiments setup}

\paragraph{Datasets.}
We pre-train our generalizable model using DTU~\cite{jensen2014large} dataset, and take the train-test split and evaluation setting in MVSNeRF, where they select 16 views as seen views and 4 remaining views as novel views for evaluation on each test scene. 
To further show the generalization ability of our method, we also test the model (trained on DTU) on the NeRF Synthetic and Real Forward-facing~\cite{mildenhall2020nerf} datasets. They both include 8 complex scenes that have different view distributions from DTU.  
For dynamic scenes, we evaluate the proposed method on the ZJU-MoCap~\cite{peng2021neural} and DynamicCap~\cite{habermann2021} datasets. ZJU-MoCap and DynamicCap datasets provide synchronized and calibrated multi-view videos with simple background and high quality masks.  
For dynamic scenes, we uniformly sample half views as seen views and the other half for testing
, and the image resolution is set to $512 \times 512$. 

\begin{table}
\tabcolsep=0.1cm
\begin{center}
    \caption{\textbf{Quantitative results on dynamic scenes.} The time within parentheses represents the training time of the model on the scene. "0" means that we directly apply the model to the scene without additional fine-tuning. All methods are evaluated on the same machine to report the speed of rendering a $512 \times 512$ image. NB denotes NeuralBody.}
\label{table:mask}
\resizebox{\columnwidth}{!}{
\begin{tabular}{l|ccc|ccc|c}

\Xhline{3\arrayrulewidth}
 & \multicolumn{3}{c|}{ZJU-MoCap} & \multicolumn{3}{c|}{DynamicCap}  \\ 
\hline
  & PSNR  & SSIM & LPIPS  & PSNR  & SSIM  & LPIPS  & FPS  \\
\hline
NB (10h) & 31.85 & 0.971 & 0.079 & 25.35 &  0.908 & 0.153 & 1.297\\
DNeRF (10h) & 26.53 & 0.889 & 0.154 & 22.52 & 0.826 & 0.258 & 0.079\\
IBRNet (0) & 29.46 & 0.947 &  0.094 & 24.67 & 0.906 & 0.130 & 0.517 \\
IBRNet (2h) & 32.38 & 0.968 & 0.065 & 25.38 & 0.900 & 0.137 & 0.517 \\
\hline
Ours (0) & 31.21 & 0.970 & 0.041 & 26.29 & 0.941 & 0.068  & \textbf{40.21} \\
Ours (15min) & 32.12 & 0.972 & 0.037 & 26.80 & 0.941 &  0.065 & \textbf{40.21} \\
Ours (2h) & \textbf{32.52} & \textbf{0.978} & \textbf{0.030} & \textbf{27.07} & \textbf{0.944} & \textbf{0.059} &  \textbf{40.21} \\
\Xhline{3\arrayrulewidth}
\end{tabular}
}
\end{center}


\end{table}

\paragraph{Baselines.}
As a generalizable radiance field method, we first make comparisons with PixelNeRF~\cite{yu2020pixelnerf}, IBRNet~\cite{wang2021ibrnet} and MVSNeRF~\cite{chen2021mvsnerf}, which are recent state-of-the-art open source radiance field methods. 
Then we show that the proposed method with a short fine-tuning process can achieve comparable results with NeRF and other per-scene optimization methods~\cite{wang2021ibrnet, chen2021mvsnerf}.
We follow the same setting as MVSNeRF on above benchmarks and borrow the results of baselines from MVSNeRF.
On dynamic scenes, we make comparisons with deformation fields based methods \cite{pumarola2020d, zhang2021editable} and generalizable radiance field methods~\cite{wang2021ibrnet}. We also compare our method with NeuralBody~\cite{peng2021neural}.
For DNeRF~\cite{pumarola2020d} and NeuralBody, we use their released code and retrained their method under our experimental setting.
For IBRNet~\cite{wang2021ibrnet}, we use their released code and evaluate their released model pretrained on multiple large-scale datasets under our setting.

\begin{figure*}
    \centering
    \includegraphics[width=0.94\textwidth]{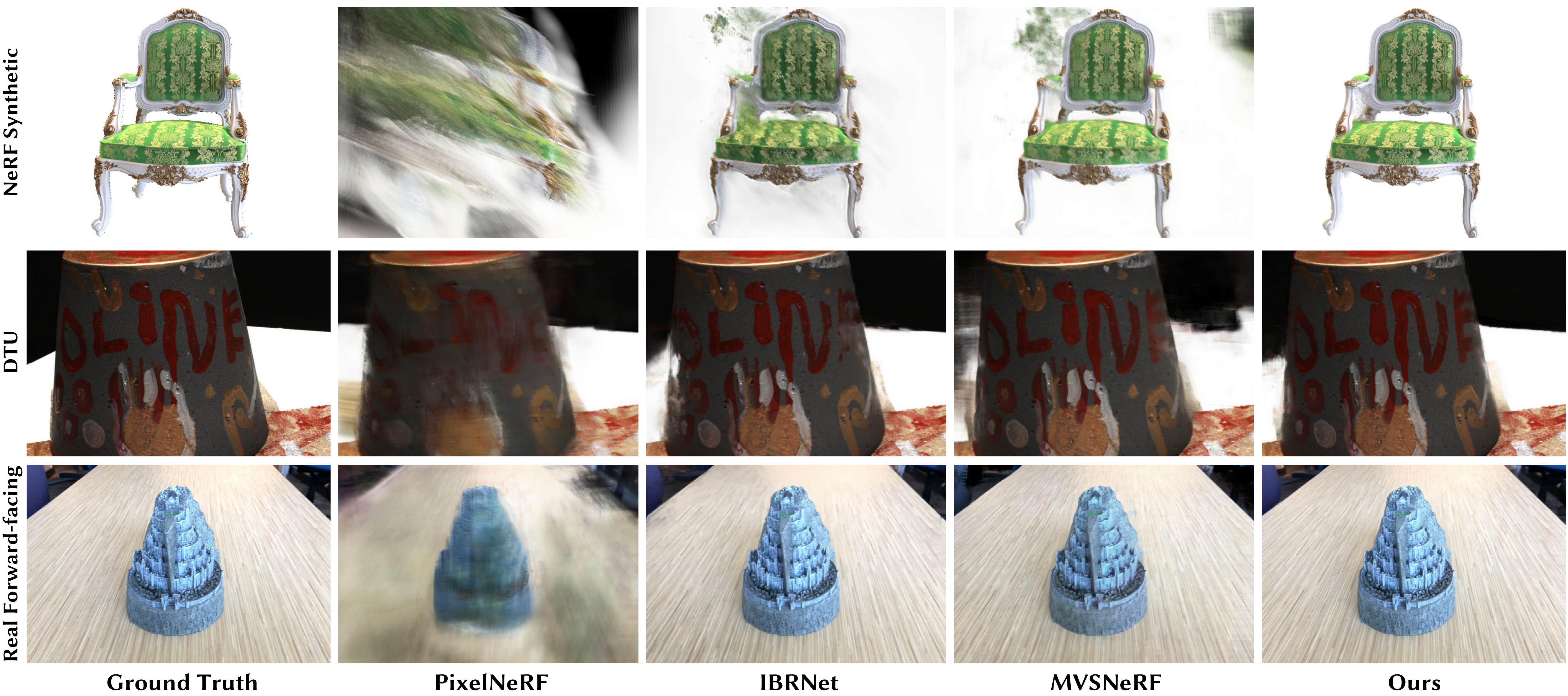}
    \caption{\textbf{Qualitative comparison of image synthesis results under the generalization setting.}}
    \label{fig:comp_gen}
\end{figure*}

\begin{table*}

  \centering
  \caption{\textbf{Quantitative results of static scenes.} ``Generalization'' means that the model is not additionally fine-tuned. ``Per-scene optimization'' means that the model is trained or fine-tuned on the target scene. All methods are evaluated on an RTX 3090 GPU to report the speed of rendering a 512 $\times$ 512 image. The results of baselines are borrowed from MVSNeRF.}
  \label{tab:comp_syn_dtu_llff}
  \resizebox{0.9\linewidth}{!}{
  \begin{tabular}{l|c|r|ccc|ccc|ccc}
    \Xhline{3\arrayrulewidth}
    \multirow{2}{*}{Methods} & \multirow{2}{*}{\begin{tabular}[c]{@{}c@{}}Training\\ settings \end{tabular}} & \multirow{2}{*}{\begin{tabular}[c]{@{}c@{}} FPS $\uparrow$ \end{tabular}} &\multicolumn{3}{c|}{NeRF Synthetic} & \multicolumn{3}{c|}{DTU} & \multicolumn{3}{c}{Real Forward-facing}\\ \cline{4-12} 
         & & & PSNR$\uparrow$ & SSIM$\uparrow$ & LPIPS$\downarrow$ & PSNR$\uparrow$ & SSIM$\uparrow$ & LPIPS$\downarrow$ & PSNR$\uparrow$ & SSIM$\uparrow$ & LPIPS$\downarrow$ \\
    \hline
     PixelNeRF &  & 0.019 & 7.39 & 0.658 & 0.411 & 19.31 & 0.789 & 0.382 & 11.24 & 0.486 & 0.671  \\ 
    IBRNet &  \multirow{2}{*}{\begin{tabular}[c]{@{}l@{}} Generalization \end{tabular}}
    & 0.217 & 22.44 & 0.874 & 0.195 & 26.04 & 0.917 & 0.191 & 21.79 & 0.786 & 0.279 \\
    MVSNeRF &  & 0.416 & 23.62 & 0.897 & 0.176 & 26.63 & 0.931 & 0.168 & 21.93 & 0.795 & 0.252 \\
    Ours &  & 25.29 & \textbf{26.65} & \textbf{0.947} & \textbf{0.072} & \textbf{27.61} & \textbf{0.956} & \textbf{0.091} & \textbf{22.78} & \textbf{0.808} & \textbf{0.209} \\
    \hline
    NeRF$_{10.2h}$ &  & 0.151 & \textbf{30.63} & \textbf{0.962} & 0.093 & 27.01 & 0.902 & 0.263 & \textbf{25.97} & 0.870 & 0.236  \\
    IBRNet$_{ft-1.0h}$ & \multirow{2}{*}{\begin{tabular}[c]{@{}l@{}} Per-scene \\ optimization \end{tabular}} & 0.217 & 25.62 & 0.939 & 0.111 & \textbf{31.35} & 0.956 & 0.131 & 24.88 & 0.861 & 0.189 \\
    MVSNeRF$_{ft-15min}$ & & 0.416 & 27.07 & 0.931 & 0.168 & 28.51 & 0.933 & 0.179 & 25.45 & \textbf{0.877} & 0.192 \\
    Ours$_{ft-15min}$ &  & 25.29 & 27.20 & 0.951 & 0.066 & 28.73 & 0.956 & 0.093 & 24.59 & 0.857 & 0.173 \\
    Ours$_{ft-1.0h}$ &  & 25.29 & 27.57 & 0.954 & \textbf{0.063} & 28.87 & \textbf{0.957} & \textbf{0.090} & 24.89 & 0.865 & \textbf{0.159}\\
    \Xhline{3\arrayrulewidth}
  \end{tabular}
  }
  \vspace{-3mm}

\end{table*}

\subsection{Performance on image synthesis}

\paragraph{Comparisons on dynamic scenes.}
Table~\ref{table:mask} lists the quantitative results, which show that our method achieves the state-of-the-art performance with real-time rendering speed. Fig.~\ref{fig:dynamiccap} presents the qualitative results. Note that the length of selected sequences on these datasets are generally from 600 to 1000 frames. We find that DNeRF~\cite{pumarola2020d} has difficulty in handling theses scenes because long sequences have very complex motions. 
To help DNeRF handle the complex dynamic scenes, we divide the video sequence into several sub-sequences, each of which is separately modeled by a DNeRF network. Please refer to the supplementary material for more details of the setup for baselines.

\begin{figure*}
    \centering
    \includegraphics[width=0.93\textwidth]{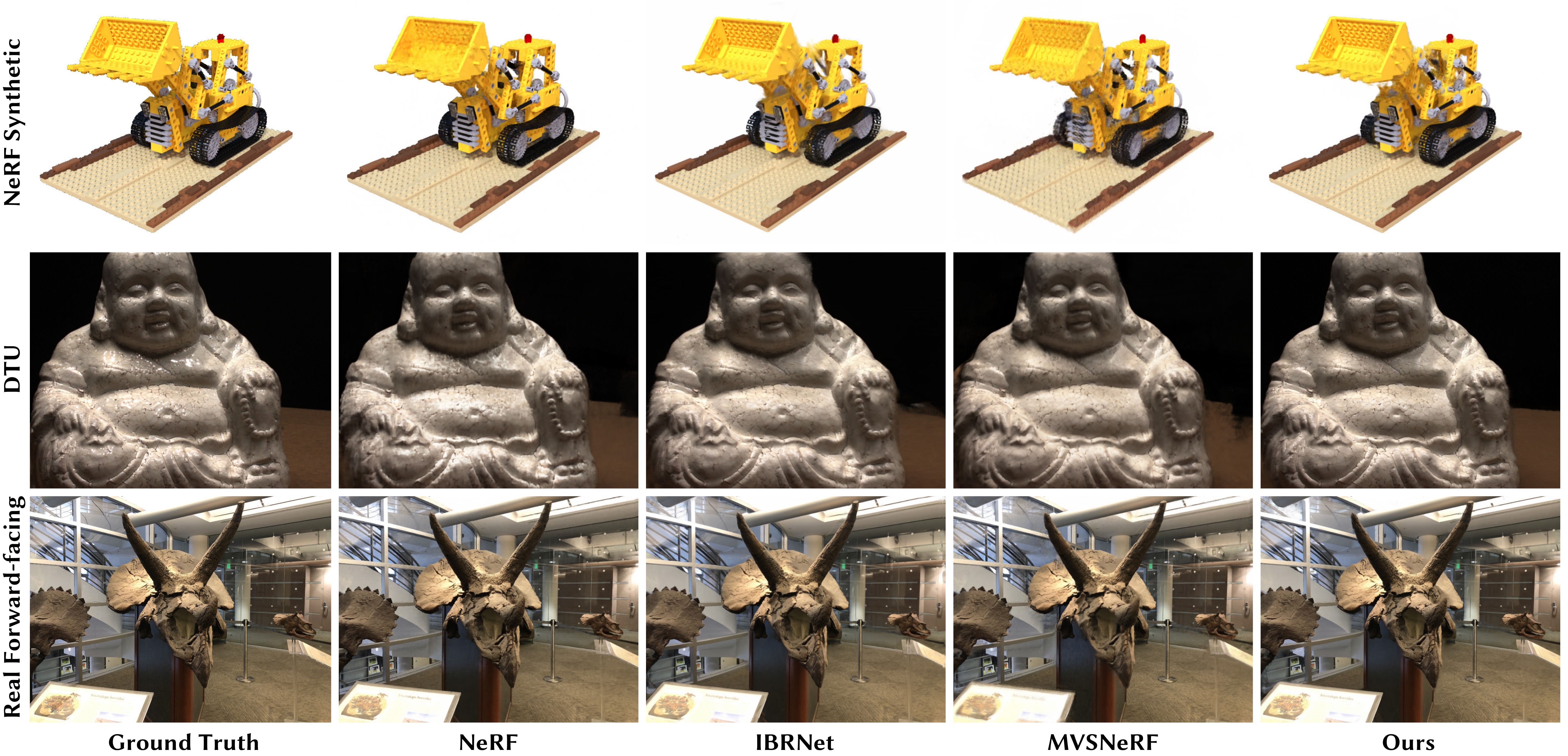}    
    \caption{\textbf{Qualitative comparison of image synthesis results under the per-scene optimization setting.}}
    \label{fig:comp_ft}
\end{figure*}

\paragraph{Comparisons on static scenes.}
We report metrics of PSNR, SSIM \cite{wang2004image} and LPIPS \cite{zhang2018unreasonable} on the NeRF Synthetic, DTU and Real Forward-facing datasets in Table~\ref{tab:comp_syn_dtu_llff}. 
As shown by the quantitative results, our method exhibits competitive performance with baselines with a significant gain in the rendering speed.
Note that since PixelNeRF uses a much wider MLP than other methods, it takes almost 10x as much time for rendering compared to other baselines.
Fig.~\ref{fig:comp_gen} and \ref{fig:comp_ft} present the qualitative results.

\begin{table}
  \tabcolsep=0.04cm
  \begin{center}
  \caption{\textbf{Quantitative comparison of depth results on the DTU dataset.}  We follow the experimental setting of MVSNeRF and borrow the baseline results from it. MVSNet is trained with depth supervision while other methods are trained with RGB supervision. Abs err means the absolute error. Acc(X) metric is the percentage of pixels whose error is less than X mm.}
  \label{tab:comp_depth}
  \resizebox{\columnwidth}{!}{
  \begin{tabular}{l|ccc|ccc}
  \Xhline{3\arrayrulewidth}
  & \multicolumn{3}{c|}{Reference view} & \multicolumn{3}{c}{Novel view} \\
  \hline
    & Abs err & Acc(2) & Acc(10) & Abs err & Acc(2) & Acc(10) \\
  \hline
  MVSNet & \textbf{3.60} & 0.603 & \textbf{0.955} & - & - & - \\ 
  PixelNeRF & 49 & 0.037 & 0.176 & 47.8 & 0.039 & 0.187 \\
  IBRNet & 338 & 0.000  & 0.913 & 324 & 0.000 & 0.866\\
  MVSNeRF & 4.60 &  0.746& 0.913  & 7.00 & 0.717 & 0.866\\
  Ours-MVS & 3.80 & 0.823 & 0.937 & 4.80 & 0.778 & 0.915 \\
  Ours-NeRF & 3.80 & \textbf{0.837} & 0.939 & \textbf{4.60} & \textbf{0.792} & \textbf{0.917}  \\
  \Xhline{3\arrayrulewidth}
  \end{tabular}
  }
  \end{center}
  
  \vspace{-1em}
  
  \end{table}

\begin{table}
\centering
\tabcolsep=0.15cm
\caption{\textbf{Quantitative ablation of the design choices on the DTU dataset.} ``Depth-gui.'' and ``Depth-sup.'' are ``Depth-guided'' and ``Depth-supervision'', respectively. The rendering resolution here is set to $512 \times 640$. }
\resizebox{\columnwidth}{!}{
\begin{tabular}{cccccc}
    \Xhline{3\arrayrulewidth}
    Samples &  Depth-gui. & Cascade &  Depth-sup. & PSNR$\uparrow$ & FPS$\uparrow$  \\
    \hline
    128 & & & & 27.39 & 0.630 \\
    2   & & & & 17.75 & 11.50 \\
    2   & \cmark & & & 26.97 & 9.749\\
    2   & \cmark & \cmark &  & 27.45 & 20.31 \\
    2   & \cmark & \cmark  & \cmark & 27.11 & 20.31\\
    \Xhline{3\arrayrulewidth}
\end{tabular}}
\label{tab:ablation_main}
\end{table}

    
    

\subsection{Quality of reconstructed depth}
%
To evaluate the performance of depth prediction, we compare our method with generalizable radiance field methods~\cite{yu2020pixelnerf, wang2021ibrnet, chen2021mvsnerf} and the classic MVS method MVSNet~\cite{yao2018mvsnet} on the DTU dataset. Denote the depth reconstructed from volume densities as ``Ours-NeRF'' and the depth produced by the fine-level depth probability volume as ``Ours-MVS''.
Note that MVSNet is trained with depth supervision while other methods are trained with image supervision.
As shown in Table~\ref{tab:comp_depth}, ``Ours-NeRF'' significantly outperforms baseline methods.
Moreover, ``Ours-MVS'' also produces reasonable depth results from the cost volume.
The results show that the proposed sampling strategy facilitate our model to reconstruct better scene geometry than \cite{wang2021ibrnet, chen2021mvsnerf}.
Please refer to the supplementary material for visual results.

\subsection{Ablation studies and analysis}
\paragraph{Ablation of main proposed components.}
We firstly execute experiments to show the power of depth-guided sampling.
As shown in the first three rows of Table~\ref{tab:ablation_main}, when we reduce the number of samples from 128 to 2, the rendering quality of the method without depth-guided sampling drops a lot, 
while the method with depth-guided sampling can maintain almost the same performance.
Secondly, we conduct experiments to analyze the effect of the cascade cost volume. As shown in the fourth row of Table~\ref{tab:ablation_main}, the cascade design greatly improves the rendering speed and shows the same good performance.
Note that the non-cascade design has 128 depth planes, while the cascade design has 64 and 8 depth planes.
Finally, we additionally supervise the depth probability volume using ground truth depth. As shown in the last row of Table~\ref{tab:ablation_main}, using depth supervision does not improve the rendering performance. 




\paragraph{Running time analysis.} To render a $512 \times 512$ image, our method with 3 input views and 2 samples per ray runs at 25.21 FPS on a desktop with an RTX 3090 GPU.
Specifically, our method takes 4.3 ms to extract image features of 3 input images, 16.1 ms to process the cost volume using 3D CNN, and 19.2 ms for inferring radiance fields and volume rendering.

\section{Conclusion and discussion}
\label{sec:conclusion}

This paper introduced ENeRF which could support interactive free-viewpoint videos.
The core innovation is utilizing explicit depth maps as coarse scene geometry to guide the rendering process of implicit radiance fields.
We demonstrated competitive performance of our method among baselines while being significantly faster than previous generalizable radiance field methods.

Although our method can efficiently generate high-quality images, it still has the following limitations. 
1) This work focuses on solid surfaces and cannot handle scenes that have multiple surfaces contributing to the appearance, such as transparent scenes.
2) The proposed model generates novel views based on nearby images. Once some target regions under the novel view are invisible in input views, the rendering quality may degrade. It could be solved by considering the visibility of input views.


%

\begin{acks}
The authors would like to acknowledge support from NSFC (No. 62172364), Information Technology Center and State Key Lab of CAD\&CG ,ZheJiang University.
\end{acks}

\bibliographystyle{ACM-Reference-Format}
\bibliography{sample-bibliography}

\newpage
\;
\newpage

\setcounter{section}{0}
\section*{Supplementary material}
In the supplementary material, we provide network architectures, details of experimental setup, and more experimental results.

\section{Method details}
\subsection{Network architectures}
\paragraph{Pooling operator.}  
Given the multi-view point features $\{f_i\}_{i=1}^N$, the pooling operator $\psi$ aims to aggregate these features to obtain the feature $f_{\text{img}}$, which is used to infer the radiance field. 
Instead of simply concatenating these features like MVSNeRF~\cite{chen2021mvsnerf}, we use a weighted pooling operator proposed in IBRNet~\cite{wang2021ibrnet}, which allows us to input any number of source views.
Specifically, we first compute a per-element mean $\boldsymbol{\mu}$ and variance $\mathbf{v}$ of $\{f_i\}_{i=1}^N$ to capture global information.
Then we concatenate each feature $f_i$ with $\boldsymbol{\mu}$ and $\mathbf{v}$, and feed the concatenated feature into a small shared MLP to obtain a weight $w_i$.
The feature $f_{\text{img}}$ is blended via a soft-argmax operator using weights $\{w_i\}_{i=1}^N$ and multi-view features $\{f_i\}_{i=1}^N$.

\paragraph{Architectures of MLPs.}
The MLP $\phi$ is used to infer the density $\sigma$ from the image feature $f_{\text{img}}$ and the voxel feature $f_{\text{voxel}}$. 
To predict the color of the point, we use the MLP $\varphi$ to yield the blending weights for image colors in the source views.
We illustrate the architectures of $\phi$ and $\varphi$ in Table~\ref{tab:network}.

\subsection{Object-compositional representation}
\label{sec:object_compositional}

\begin{figure*}
    \centering
    \includegraphics[width=\linewidth]{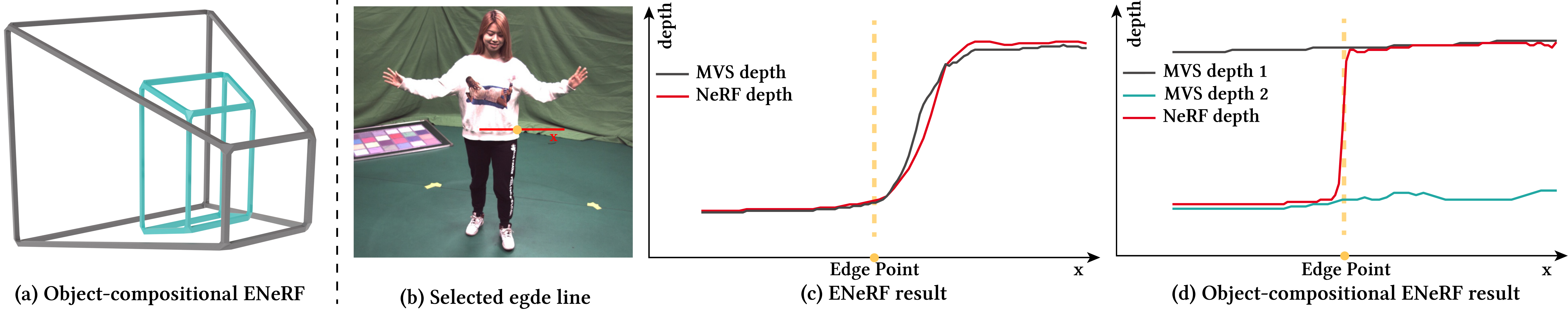}
    \caption{\textbf{Illustration of compositional ENeRF.} (a) To implement compositional ENeRF, we estimate 3D bounding boxes of foreground and background entities, and use ENeRF to separately predict their radiance fields, which are then composited to the full scene's radiance field. (b) MVS-based depth estimator tends to give smooth predictions that are inaccurate on the edge between the foreground and background. (c) By separately predicting the depth of foreground and background entities, we can sample points around accurate surfaces, thereby enabling NeRF to represent the sharp depth discontinuities.}
    \label{fig:depth_uncon}
\end{figure*}



On complex dynamic scenes (ST-NeRF dataset and outdoor datasets we collected), foreground dynamic objects are usually far away from the background, as shown in Fig.~\ref{fig:depth_uncon} (b).
We observe that ENeRF tends to produce artifacts on the border between the foreground objects and the background.
The reason is that there exists the depth discontinuities on these regions and our depth estimator tends to produce smooth predictions, resulting in inaccurate depth predictions on edge regions, as shown in Fig.~\ref{fig:depth_uncon} (c). 
To solve this problem, we introduce the object-compositional representation.
Fig.~\ref{fig:depth_uncon}~(a) presents an example.
Specifically, we first estimate 3D bounding boxes of foreground objects.
Then, ENeRF is separately applied to these foreground regions.
For simplicity, we assume that the background is static and multi-view background images are given, but our method could also be extended to handle dynamic background. We can use the depth estimator to predict the background depth using the background images.
Since the depth estimator of ENeRF is used to predict the depth of foreground and background entities respectively, we can obtain accurate scene geometry and infer radiance fields around correct scene surfaces.
Finally, the radiance fields of foreground and background entities are composited.
This representation can represent sharp discontinuities and produce accurate depth predictions on the foreground object edge regions, as shown in Fig.~\ref{fig:depth_uncon} (d).

\section{Details of the experimental setup}
\paragraph{Experimental setup on dynamic scenes.}
\label{sec:exp_dynamic}
The evaluation setup on dynamic scenes is taken from NeuralBody~\cite{peng2021neural} and is described as the following. 
Given the 3D bounding box the dynamic entity, we project it to obtain a bound\_mask and make the colors of pixels outside the mask as zero.
We compute the PSNR metric for pixels belonging to the bound\_mask.
To report the SSIM and LPIPS metrics, we first compute the 2D box that bounds the bound\_mask and then evaluate the corresponding image region.
On the ZJU-MoCap~\cite{peng2021neural} dataset, we select the first 600 frames of Sequence-313. 
On the DynamicCap~\cite{habermann2021} dataset, we select the first 1000 frames for Sequence-olek and 600 frames for Sequence-vlad.
Note that we do not make comparisons with MVSNeRF~\cite{chen2021mvsnerf} on dynamic scenes because MVSNeRF use the feature volume when finetuning, which prevent it from handling dynamic scenes.

\paragraph{Experimental setup on static scenes.}
Our evaluation setup is taken from MVSNeRF~\cite{chen2021mvsnerf} and is described as the following. 
To report the results on the DTU~\cite{jensen2014large} dataset, we compute the metric score of foreground part in images. 
For metrics of SSIM and LPIPS, we set the background to black and calculate the metric score of the whole image.
The segmentation mask is defined by whether there is ground-truth depth available at each pixel.
Since marginal regions of images are usually invisible to input images on the Real Forward-facing~\cite{mildenhall2020nerf} dataset, we only evaluate 80\% area in the center of images.
The image resolutions are set to $512 \times 640$, $640 \times 960$ and $800 \times 800$ on the DTU, Real forward-facing and NeRF Synthetic~\cite{mildenhall2020nerf} datasets, respectively. 

\begin{table}[]
    \centering
    \caption{\textbf{The architectures of MLPs.} We denote LR to be LinearRelu layer.  ``Chns.'' shows the number of input and output channels for each layer.}
    \label{tab:network}
    \resizebox{\columnwidth}{!}{
    \begin{tabular}{c|c|c|c|c}
        \Xhline{3\arrayrulewidth}
        MLP & Layer & Chns.  & Input & Output \\
        \hline
        \multirow{2}{*}{$\phi$} & LR$_0$ & 8 + 16 / 128 & $f_{\text{img}}$, $f_{\text{voxel}}$ & hidden feature \\
            & LR$_1$ & 128 / 64 + 1 & hidden feature & $f_p$, $\sigma$ \\   
        \hline
        \multirow{3}{*}{$\varphi$} & LR$_0$ & 64 + 16 + 4 / 128 & $f_p$, $f_i$, $\Delta \mathbf{d}_i$ & hidden feature \\ & LR$_1$ & 128 / 64 & hidden feature & hidden feature \\ & LR$_2$ & 64 / 1 & hidden feature & $w_i$ \\
        \Xhline{3\arrayrulewidth}
    \end{tabular}
    }
\end{table}

\begin{figure}
    \centering
    \includegraphics[width=\columnwidth]{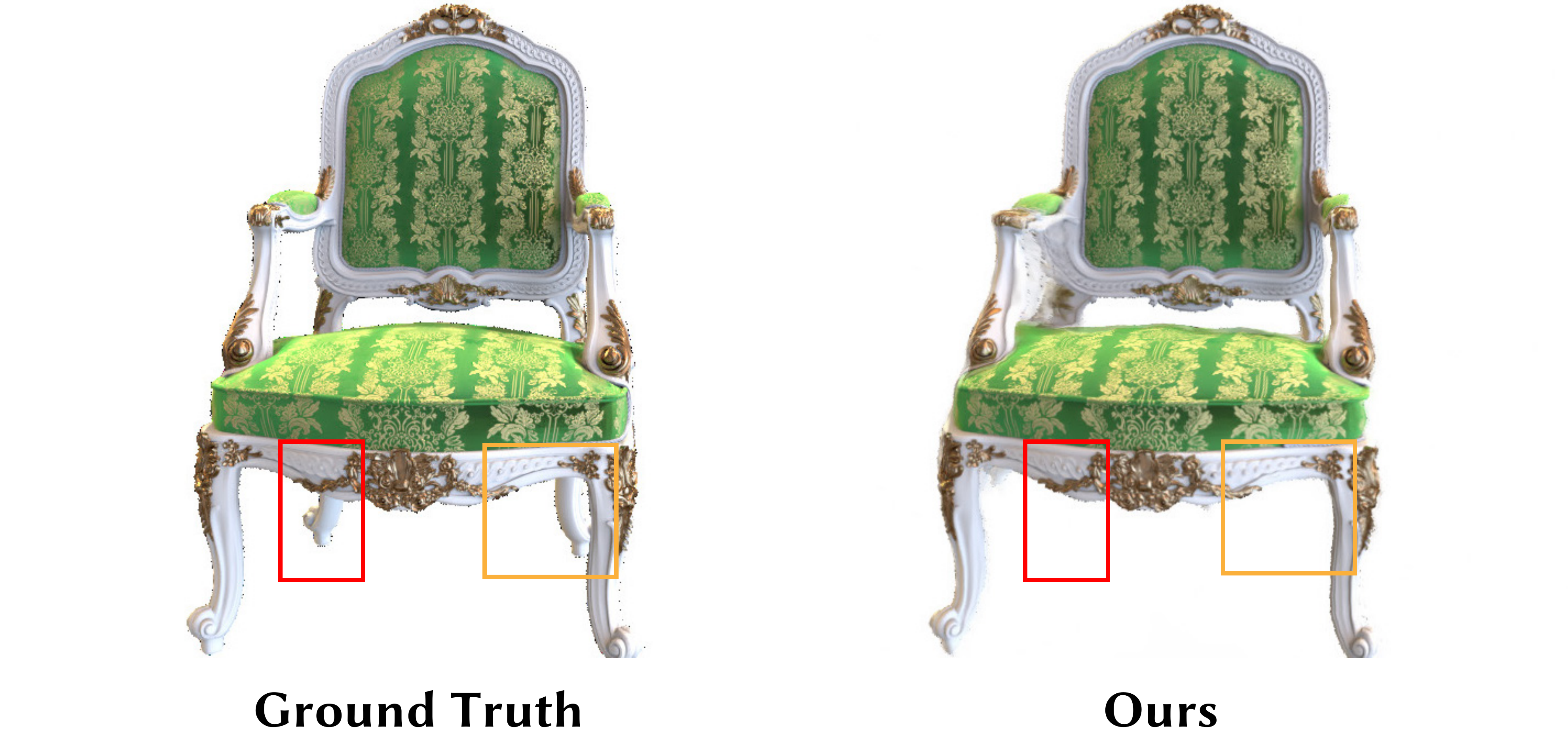}
    \caption{\textbf{Visual results of artifacts caused by occlusions.} In this case, there are some artifacts in the chair legs due to the occlusion.}
    \label{fig:occlusion}
	\vspace{-2.5mm}
\end{figure}

\paragraph{Experimental setup on the rendering FPS}
We report the rendering FPS using the same desktop with an RTX 3090 GPU.
The rendering time is defined as network forwarding time. All baselines are implemented using Pytorch.
For dynamic scenes, we could further reduce the rendering time by only rendering the image regions in the bound\_mask defined in the last section (Experimental setup on dynamic scenes).
This improves the rendering FPS from 30.57 to 40.21.
Note that all baselines on the dynamic scenes utilize this trick to ensure the comparison fairness.
The rendering FPS of free-viewpoint video demos in the video extraly includes the data loading and interactive GUI time. All processes in the interactive demos are implemented with python and the GPU utilization is generally from 70 to 80 \%. This indicates our method may achieve higher speeds with a more efficient implementation.

\paragraph{Experimental details of DNeRF}
we found simply applying DNeRF \cite{pumarola2020d} on benchmark datasets cannot converge because benchmark datasets exhibit complex motions. 
To obtain a reasonable result for DNeRF, we train 2 models for the Sequence-313 on the ZJU-MoCap dataset.
On the DynamicCap dataset, we train 10 and 12 models for Sequence-olek and Sequence-vlad, respectively.
Specifically, for the Sequence-vlad of 600 frames, training 12 models means that each model is trained on 50 frames.





\section{More experimental results}

\begin{table}
  \centering
\caption{\textbf{Quantitative ablation analysis of the number of samples and input source views on the DTU dataset.}}
  \begin{tabular}{ccc|ccc}
    \Xhline{3\arrayrulewidth}
    Samples & PSNR$\uparrow$ & FPS$\uparrow$ & Views & PSNR$\uparrow$ & FPS$\uparrow$ \\
    \hline
    1  & 26.89 & 26.01 & 2 & 25.45 & 24.91\\
    2  & 27.45 & 20.01 & 3 & 27.45 & 20.01\\
    4  & 27.49 & 13.42  & 4 & 27.80 & 16.73\\
    8  & 27.54 & 8.09  & 5 & 27.84 & 14.32 \\
    \Xhline{3\arrayrulewidth}
  \end{tabular}
\label{tab:ablation_samples_views}
\end{table}

\begin{figure}
    \centering
    \includegraphics[width=\columnwidth]{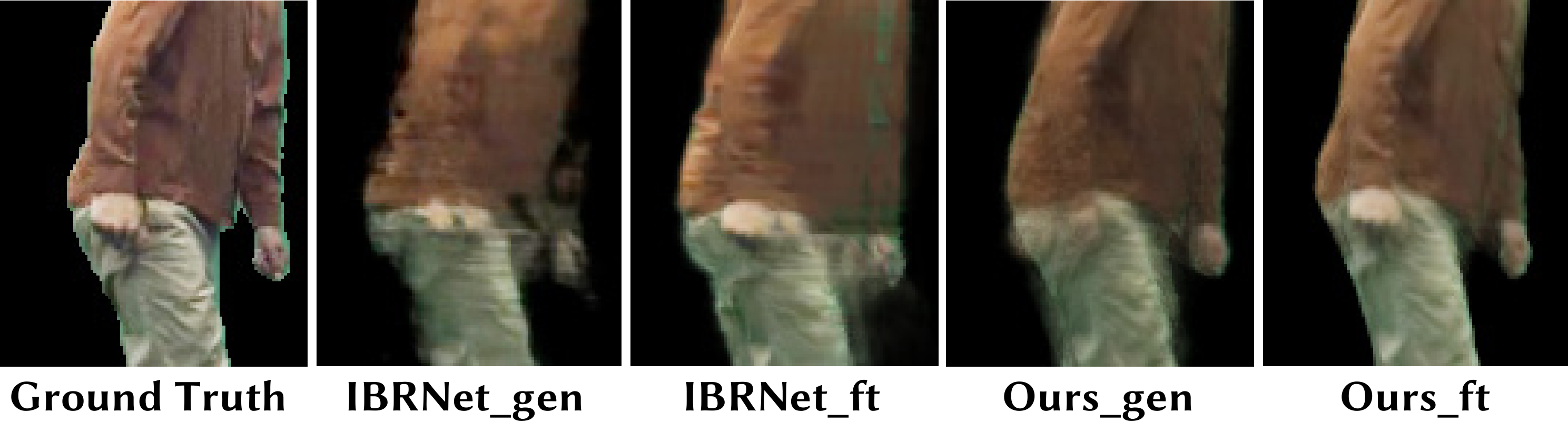}
    \caption{\textbf{Zoom-in results on the DynamicCap dataset.} Our method renders more appearance details than baseline methods. After short-term finetuning, the proposed method produces better rendering results.}
    \label{fig:zoomin}
	\vspace{-4mm}
\end{figure}

\paragraph{Visual results of artifacts caused by occlusions.} As discussed in section 5, our method may have artifacts in occluded regions. We provide visual results on the NeRF Synthetic dataset under the generalization setting in Fig.~\ref{fig:occlusion} to identify these artifacts.

\begin{figure*}[t]
    \centering
    \includegraphics[width=0.99\textwidth]{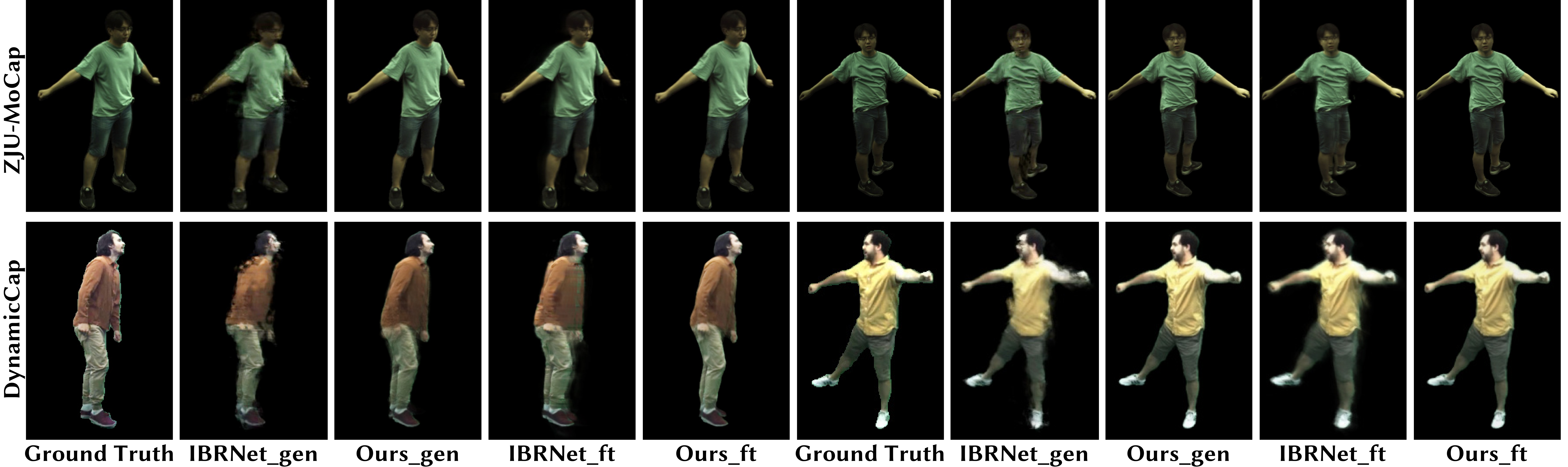}
    \caption{\textbf{Quantitative results on the ZJU-MoCap and DynamicCap datasets.} ``IBRNet\_gen'' and ``Ours\_gen'' mean that these models are directly applied to input images without be additionally fine-tuned on input videos. ``IBRNet\_ft'' and ``Ours\_ft'' mean that these models are finetuned on the target sequence.}
    \label{fig:dynamiccap_supp}
\end{figure*}

\paragraph{Visual results of fine-tuning models on dynamic scenes.}
We include the visual results of fine-tuning models on ZJU-MoCap and DynamicCap datasets in Fig.~\ref{fig:dynamiccap_supp} and zoom-in results in Fig.~\ref{fig:zoomin}. Our method exhibits better performance after fine-tuning on the target sequence.

\begin{figure*}
    \centering
    \includegraphics[width=\linewidth]{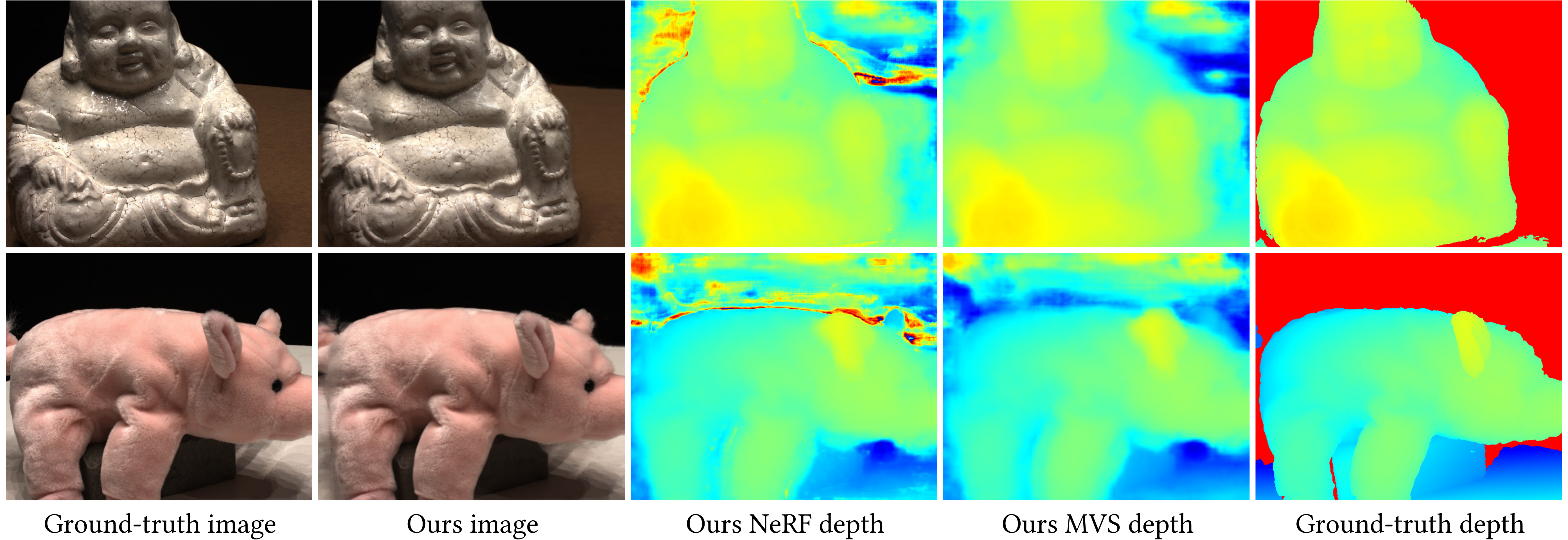}
    \caption{\textbf{Visual depth results on the DTU dataset. }``Ours NeRF depth'' represents the depth results recovered from volume densities. ``Ours MVS depth'' denotes the depth results from the fine-level cost volume.}
    \label{fig:depth}
\end{figure*}

\paragraph{Visual depth results.} 
As shown in Figure~\ref{fig:depth}, the proposed method produces reasonable depth results by supervising networks with only RGB images.
The cost volume recovers high-quality depth, which allows us to place few samples around the surface to achieve photorealistic view synthesis.

\begin{figure*}[]
    \centering
    \includegraphics[width=\linewidth]{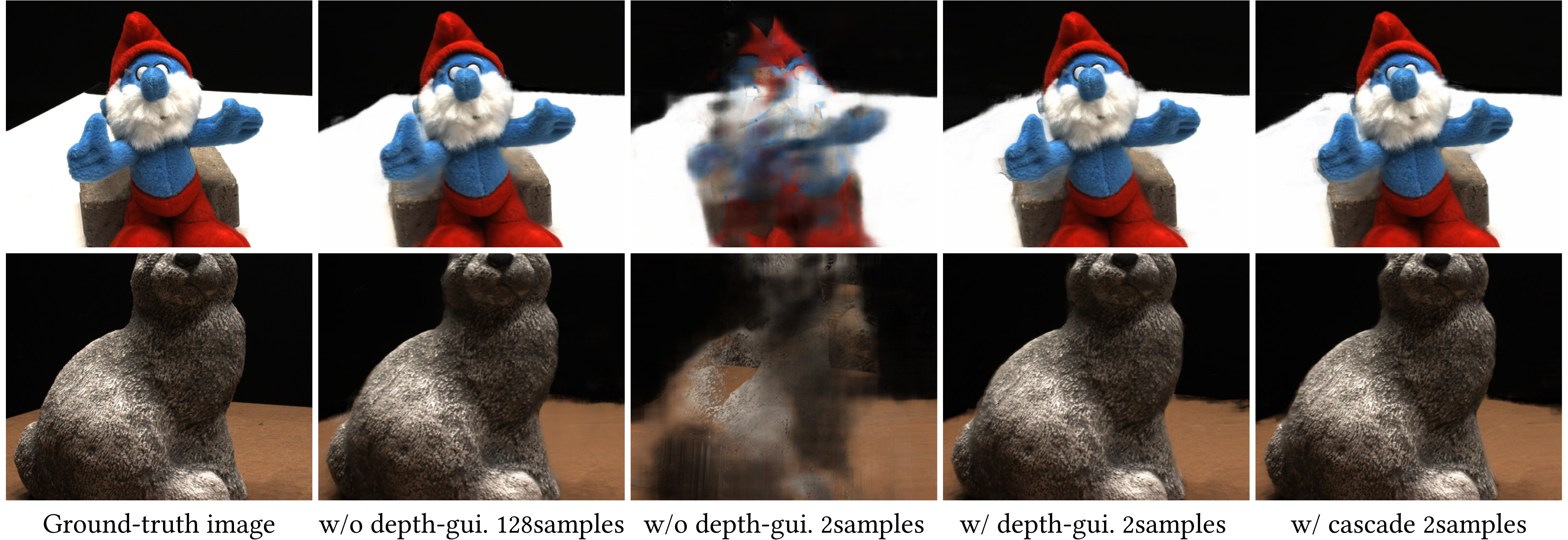}
    \caption{\textbf{Visual ablation results on the DTU dataset.} ``w/o depth-gui.'' is similar to MVSNeRF~\cite{chen2021mvsnerf}. }
    \label{fig:ablation}
\end{figure*}

\paragraph{Visual ablation results.}
We provide visual ablation results in Figure~\ref{fig:ablation}. 
The results show that when we reduce the number of samples from 128 to 2, our method with depth-guided sampling almost maintains the same rendering quality.
With the depth-guided sampling, the construction of a high-resolution cost volume becomes a bottleneck in the rendering speed.
The cascade cost volume further speeds up the construction of the cost volume without loss of rendering quality as shown in Figure~\ref{fig:ablation}.

\paragraph{Sensitivity analysis.} We provide sensitivity analysis on the number of input views and the number of samples per ray in Table~\ref{tab:ablation_samples_views}. On the DTU dataset, we found 3 views and 2 sampling points can produce  high quality view synthesis results at interactive frame rates.

\paragraph{Ablation on the rendering head.} As stated in the main paper, we predict the blend weights for source view RGBs to obtain the final radiance like~\cite{wang2021ibrnet}. Alternatively, we can also output the final radiance simply by predicting the radiance from the volume and image feature like~\cite{chen2021mvsnerf}. The results are 26.77/0.95 in terms of PSNR/SSIM metrics when using the rendering head in ~\cite{chen2021mvsnerf}, while they are 27.88/0.96 when using the rendering head in ~\cite{wang2021ibrnet}.

\paragraph{Ablation on the depth range hyperparameter $\lambda$.} 
We experimentally found that setting $\lambda = 1$ produces good results.
One optional design maybe progressively reduce the $\lambda$ from 3 to 1 during training. This design produces lower performance than the taken design (PSNR/SSIM = 25.25/0.89 v.s. 27.61/0.96).

\paragraph{Perceptual loss.} We find that the perceptual loss largely affects the LPIPS metric but has little effect on PSNR and SSIM metrics. Specifically, the model trained with $\mathcal{L}_{perc}$ gives 27.62/0.956/0.091 on the DTU dataset in terms of PSNR/SSIM/LPIPS metrics, while the one without $\mathcal{L}_{perc}$ gives 27.88/0.955/0.106.

\paragraph{Comparisons with NeuralRays~\cite{liu2021neural}}. We provide qualitative comparisons with NeuralRays under the experimental setting taken in their paper where various large-scale datasets are used during pretraining.
Our method presents competitive performance while runs significantly faster than their method on the Real Forward-facing dataset.(Ours:25.71/8.42 v.s. NeuralRays:25.35/0.03 in terms of PSNR/FPS@756x1008; the performance is from their paper and the speed is tested using their released code).

\begin{table}
	\centering
	\caption{\textbf{Quantitative comparison on the DTU dataset.}}
	\begin{tabular}{l|ccccc}
		\Xhline{3\arrayrulewidth}
        Scan &  \#1 & \#8 & \#21 & \#103 & \#114 \\ \hline
		Metric & \multicolumn{5}{c}{PSNR$\uparrow$}\\ \hline
		PixelNeRF & 21.64&  23.70& 16.04& 16.76&18.40\\
		IBRNet & 25.97&  27.45 & 20.94& 27.91&27.91\\
		MVSNeRF & 26.96 & 27.43 & 21.55 &29.25 &27.99\\ 
		Ours & \textbf{28.85} & \textbf{29.05} & \textbf{22.53} & \textbf{30.51} & \textbf{28.86} \\
		\hline
		NeRF$_{10.2h}$ & 26.62& 28.33 & 23.24 & 30.40 &26.47\\
		IBRNet$_{ft-1h}$ & \textbf{31.00} & \textbf{32.46} & \textbf{27.88} & \textbf{34.40}& \textbf{31.00} \\
		MVSNeRF$_{ft-15min}$&  28.05 & 28.88 & 24.87 & 32.23 & 28.47 \\ 
		Ours$_{ft-15min}$ & 29.81 & 30.06 & 22.50 & 31.57 & 29.72 \\
		Ours$_{ft-1h}$ & 30.10 & 30.50 & 22.46 & 31.42 & 29.87 \\
		\hline

		Metric & \multicolumn{5}{c}{SSIM$\uparrow$}\\ \hline
		PixelNeRF &  0.827& 0.829&  0.691& 0.836&0.763\\
		IBRNet &  0.918& 0.903&  0.873& 0.950&0.943\\
		MVSNeRF & 0.937 & 0.922 & 0.890 & 0.962 & 0.949\\ 
		Ours & \textbf{0.958} & \textbf{0.955} & \textbf{0.916} & \textbf{0.968} & \textbf{0.961} \\ \hline
		NeRF$_{10.2h}$ &  0.902 & 0.876 &  0.874 & 0.944 & 0.913\\
		IBRNet$_{ft-1h}$ &  0.955& 0.945&  \textbf{0.947}& 0.968&0.964\\
		MVSNeRF$_{ft-15min}$& 0.934 & 0.900& 0.922&0.964 & 0.945\\
		Ours$_{ft-15min}$ & 0.964 & 0.958 & 0.922 & 0.971 & 0.965 \\
        Ours$_{ft-1h}$ & \textbf{0.966} & \textbf{0.959} & 0.924 & \textbf{0.971} & \textbf{0.965} \\\hline

		Metric & \multicolumn{5}{c}{LPIPS $\downarrow$ }\\ \hline

		PixelNeRF & 0.373&  0.384& 0.407& 0.376&0.372\\
		IBRNet &  0.190&   0.252& 0.179&  0.195 &  0.136\\
		MVSNeRF & 0.155 & 0.220 &0.166&0.165& 0.135\\
		Ours & \textbf{0.086} & \textbf{0.119} & \textbf{0.107} & \textbf{0.107} & \textbf{0.076} \\ \hline
		NeRF$_{10.2h}$ &  0.265&  0.321&0.246& 0.256 & 0.225\\
		IBRNet$_{ft-1h}$ & 0.129 &0.170 &0.104 &0.156& 0.099 \\
		MVSNeRF$_{ft-15min}$ & 0.171 &0.261&0.142&0.170&0.153 \\
		Ours$_{ft-20min}$ & 0.074 & 0.109 & 0.100 & 0.103 & 0.075 \\
		Ours$_{ft-1h}$ & \textbf{0.071} & \textbf{0.106} & \textbf{0.097} & \textbf{0.102} & \textbf{0.074} \\
		\Xhline{3\arrayrulewidth}
\end{tabular}
\label{tab:dtu}
\end{table}





\section{Discussions}

\textbf{Learning the depth prediction with the rendering loss.} 
We found that the depth prediction network can give good results even when we only sample one point during the volume rendering.
This is interesting, because backpropagating gradient through only one point may make the training of the depth prediction easily trapped in the local optimum.
A possible reason is that the depth is estimated using the softargmax of depth probability volume and gradients are backpropagated to the depth planes during training.
This regularizes the optimization process and helps the depth prediction network to converge to the global optimum.
To validate this assumption, we design a baseline by replacing the depth prediction network in our model with an MLP network, which takes the spatial coordinate and viewing direction as input and outputs the distance to the surface.
In contrast to our model, this baseline does not converge well and obtains bad rendering results.

\paragraph{Performance of PixelNeRF \cite{yu2020pixelnerf}.} We found that PixelNeRF gives bad performance on the NeRF Synthetic and Real Forward-facing datasets.
One reason is that PixelNeRF takes absolute XYZ coordinates as input and is trained on the DTU dataset. Since the coordinate system of the DTU dataset is quite different from that of NeRF synthetic and Real Forward-facing datasets, PixelNeRF cannot generalize well on these two datasets. MVSNeRF \cite{chen2021mvsnerf} also reported that PixelNeRF performs poor on the NeRF Synthetic and Real Forward-facing datasets.
Our method takes image features and volume features as input, thereby achieving a better generalization ability.

\paragraph{More discussions on the limitations.}
In addition to the limitations we have discussed in the main paper, we here provide more discussions. 
Our method may fail to render high-quality results in scenes including multi-surface, fuzzy, specular highlight and semi-transparent regions.
This is because the MVS generally fails to recover the correct depth in these regions.
Incorporating a layered representation~\cite{zhang2021editable} maybe helpful and we leave it as the future work.

\section{Per-scene breakdown}
Tables \ref{tab:dtu}, \ref{tab:nerf} and \ref{tab:llff} present the per-scene comparisons.
These results are consistent with the averaged results in the paper and show that our method achieves comparable performance to baselines.

\begin{table*}[t]
	\centering
	\caption{\textbf{Quantitative comparison on the NeRF Synthetic dataset.}}
	\begin{tabular}{l|cccccccc}
		\Xhline{3\arrayrulewidth}
        &   Chair & Drums & Ficus & Hotdog & Lego & Materials & Mic & Ship \\ \hline
		Metric & \multicolumn{8}{c}{PSNR$\uparrow$  }\\ \hline

		PixelNeRF   &    7.18 & 8.15  & 6.61  & 6.80   & 7.74 & 7.61      & 7.71& 7.30\\
		IBRNet    &    24.20 & 18.63 & 21.59  & 27.70  & 22.01 & 20.91     & 22.10& 22.36\\
		MVSNeRF        &   23.35 & 20.71 & 21.98 & 28.44  & 23.18& 20.05 & 22.62 & 23.35\\ 
		Ours & \textbf{28.44} & \textbf{24.55} & \textbf{23.86} & \textbf{34.64} & \textbf{24.98} & \textbf{24.04} & \textbf{26.60} & \textbf{26.09} \\ \hline
		NeRF        &    \textbf{31.07}  & \textbf{25.46}  & \textbf{29.73}  & 34.63 & \textbf{32.66}    &\textbf{30.22}& \textbf{31.81}& \textbf{29.49}\\
		IBRNet$_{ft-1h}$    &    28.18 & 21.93 & 25.01  & 31.48  & 25.34 & 24.27 & 27.29& 21.48\\
		MVSNeRF$_{ft-15min}$ &  26.80 & 22.48 & 26.24 & 32.65 & 26.62 & 25.28 & 29.78 & 26.73 \\ 
		Ours$_{ft-15min}$ & 28.52 & 25.09 & 24.38 & 35.26 & 25.31 & 24.91 & 28.17 & 25.95 \\ 
		Ours$_{ft-1h}$ & 28.94 & 25.33 & 24.71 & \textbf{35.63} & 25.39 & 24.98 & 29.25 & 26.36 \\ \hline

		Metric & \multicolumn{8}{c}{SSIM$\uparrow$ }\\\hline

		PixelNeRF & 0.624 & 0.670 & 0.669 & 0.669  & 0.671& 0.644   & 0.729& 0.584\\
		IBRNet   &    0.888& 0.836 & 0.881  & 0.923  & 0.874 & 0.872     & 0.927&  0.794\\
		MVSNeRF     &   0.876 & 0.886 & 0.898 & 0.962 & 0.902 & 0.893 & 0.923 & 0.886 \\ 
		Ours & \textbf{0.966} & \textbf{0.953} & \textbf{0.931} & \textbf{0.982} & \textbf{0.949} & \textbf{0.937} & \textbf{0.971} & \textbf{0.893} \\ \hline
		NeRF      &  0.971& 0.943& \textbf{0.969}& 0.980&\textbf{0.975}& \textbf{0.968}& 0.981 &  \textbf{0.908}\\
		IBRNet$_{ft-1h}$    & 0.955 & 0.913 & 0.940  & 0.978  & 0.940 & 0.937 & 0.974& 0.877\\
		MVSNeRF$_{ft-15min}$ &  0.934 & 0.898 & 0.944 & 0.971 & 0.924 & 0.927 & 0.970&0.879 \\ 
		Ours$_{ft-15min}$ & 0.968 & 0.958 & 0.936 & 0.984 & 0.948 & 0.946 & 0.981 & 0.891 \\
		Ours$_{ft-1h}$ & \textbf{0.971} & \textbf{0.960} & 0.939 & \textbf{0.985} & 0.949 & 0.947 & \textbf{0.985} & 0.893 \\
		\hline

		Metric & \multicolumn{8}{c}{LPIPS $\downarrow$}\\ \hline

		PixelNeRF   &  0.386 & 0.421 & 0.335 & 0.433  & 0.427& 0.432     & 0.329 & 0.526\\
		IBRNet   &    0.144 &  0.241 & 0.159 & 0.175 & 0.202 & 0.164  & 0.103 & 0.369 \\
		MVSNeRF       &  0.282& 0.187 & 0.211 & 0.173 & 0.204 & 0.216 & 0.177 & 0.244 \\ 
		Ours & \textbf{0.043} & \textbf{0.056} & \textbf{0.072} & \textbf{0.039} & \textbf{0.075} & \textbf{0.073} & \textbf{0.040} & \textbf{0.181} \\ \hline
		NeRF  & 0.055 &  0.101 & \textbf{0.047} & 0.089 &\textbf{0.054}&0.105& 0.033& 0.263\\
		IBRNet$_{ft-1h}$    & 0.079 & 0.133 & 0.082  & 0.093  & 0.105 & 0.093 & 0.040 & 0.257\\
		MVSNeRF$_{ft-15min}$ &  0.129& 0.197 & 0.171 & 0.094 & 0.176 & 0.167 & 0.117 & 0.294 \\
		Ours$_{ft-15min}$ & 0.033 & 0.047 & 0.069 & 0.031 & 0.073 & 0.063 & 0.021  & 0.190\\
		Ours$_{ft-1h}$ & \textbf{0.030} & \textbf{0.045} & 0.071 & \textbf{0.028} & 0.070 & \textbf{0.059} & \textbf{0.017}  & \textbf{0.183}\\
		\Xhline{3\arrayrulewidth}
\end{tabular}

\label{tab:nerf}
\end{table*}


\begin{table*}[b]
	\centering
	\caption{\textbf{Quantitative comparison on the Real Forward-facing dataset.}}
	\begin{tabular}{l|cccccccc}
		\Xhline{3\arrayrulewidth}
        &  Fern  &Flower&Fortress & Horns & Leaves & Orchids & Room & Trex\\ \hline        
		& \multicolumn{8}{c}{PSNR$\uparrow$  }\\ \hline

		PixelNeRF & 12.40 & 10.00 &14.07&11.07& 9.85 &9.62&11.75& 10.55\\
		IBRNet    & 20.83 & 22.38 & 27.67 & 22.06 & \textbf{18.75} & 15.29 & 27.26& 20.06\\
		MVSNeRF      & \textbf{21.15} & 24.74 & 26.03 & \textbf{23.57} & 17.51& \textbf{17.85} &26.95&\textbf{23.20}\\
		Ours & 20.88 & \textbf{24.78} & \textbf{28.63} & 23.51 & 17.78 & 17.34 & \textbf{28.94} & 20.37 \\ \hline
		NeRF$_{10.2h}$ & \textbf{23.87} & 26.84 &\textbf{31.37}&25.96& 21.21 &19.81&\textbf{33.54}& \textbf{25.19}\\
		IBRNet$_{ft-1h}$    & 22.64 & 26.55 & 30.34  & 25.01  & \textbf{22.07} & 19.01 & 31.05 & 22.34\\
		MVSNeRF$_{ft-15min}$ &  23.10 & 27.23  &30.43 & \textbf{26.35} & 21.54 & \textbf{20.51} & 30.12 & 24.32 \\ 
		Ours$_{ft-15min}$ & 21.84 & 27.46 & 29.58 & 24.97 & 20.95 & 19.17 & 29.73 & 23.01 \\
		Ours$_{ft-1h}$ & 22.08 & \textbf{27.74} & 29.58 & 25.50 & 21.26 & 19.50 & 30.07 & 23.39
		\\ \hline

		& \multicolumn{8}{c}{SSIM$\uparrow$}\\\hline

		PixelNeRF & 0.531 & 0.433 &0.674&0.516& 0.268 & 0.317 &0.691&0.458\\
		IBRNet & 0.710 & 0.854 &0.894&0.840&\textbf{0.705}&0.571&0.950&  0.768\\
		MVSNeRF & 0.638 & 0.888 &0.872&\textbf{0.868}& 0.667 &\textbf{0.657}&0.951 & \textbf{0.868}\\ 
		Ours & \textbf{0.727} & \textbf{0.890} & \textbf{0.920} & 0.866 & 0.685 & 0.637 & \textbf{0.958 }& 0.778 \\ \hline
		NeRF$_{10.2h}$ & \textbf{0.828}& 0.897 &\textbf{0.945}&0.900& 0.792  &0.721&\textbf{0.978}&\textbf{0.899}\\
		IBRNet$_{ft-1h}$    & 0.774 & 0.909 & 0.937  & 0.904  & \textbf{0.843} & 0.705 & 0.972 & 0.842\\
		MVSNeRF$_{ft-15min}$ &  0.795 & 0.912&0.943&\textbf{0.917} & 0.826 & \textbf{0.732}&0.966 &0.895 \\ 
		Ours$_{ft-15min}$ & 0.758 & 0.919 & 0.940 & 0.893 & 0.816 & 0.710 & 0.963 & 0.855 \\
		Ours$_{ft-1h}$ & 0.770 & \textbf{0.923} & 0.940 & 0.904 & 0.827 & 0.725 & 0.965 & 0.869 \\
		\hline

		& \multicolumn{8}{c}{LPIPS $\downarrow$ }\\\hline

		 &  Fern & Flower & Fortress & Horns & Leaves & Orchids & Room & Trex\\ \hline
		PixelNeRF & 0.650 & 0.708 &0.608&0.705& 0.695 &0.721&0.611& 0.667\\
		IBRNet &0.349 & 0.224 & 0.196 & 0.285 & 0.292 &0.413& 0.161 & 0.314\\ 
		MVSNeRF   &0.238 &  0.196&0.208& 0.237& 0.313 & \textbf{0.274}&0.172& \textbf{0.184}\\ 
		Ours & \textbf{0.235} & \textbf{0.168} & \textbf{0.118} & \textbf{0.200} & \textbf{0.245} & 0.308 & \textbf{0.141} & 0.259 \\ \hline
		NeRF$_{10.2h}$ &  0.291& 0.176&0.147& 0.247& 0.301& 0.321 &  0.157 &0.245\\
		IBRNet$_{ft-1h}$    & 0.266 & 0.146 & 0.133  & 0.190  & 0.180 & 0.286 & \textbf{0.089} & 0.222\\
		MVSNeRF$_{ft-15min}$ & 0.253 & 0.143 & 0.134 & 0.188 & 0.222 & 0.258 &0.149 & 0.187 \\
		Ours$_{ft-15min}$ & 0.220 & 0.130 & 0.103 & 0.177 & 0.181 & 0.266 & 0.123 & 0.183 \\
		Ours$_{ft-1h}$ & \textbf{0.197} & \textbf{0.121} & \textbf{0.101} & \textbf{0.155} & \textbf{0.168} & \textbf{0.247} & 0.113 & \textbf{0.169} 
		
		\\
		
		\Xhline{3\arrayrulewidth}
\end{tabular}

\label{tab:llff}
\end{table*}

\end{document}